\documentclass[letterpaper, 10 pt, journal, twoside]{IEEEtran}

\usepackage{hyperref}

\hypersetup{
    colorlinks=true,
    linkcolor=blue,
    filecolor=magenta,      
    urlcolor=blue,
}

\newcommand*{\change}{\textcolor{black}}

\usepackage{graphics}           
\usepackage{times}              
\usepackage{amsmath}            
\usepackage{amssymb}            
\usepackage{graphicx}
\usepackage{algorithm}
\usepackage[noend]{algpseudocode}
\usepackage{booktabs}
\usepackage{color}
\definecolor{instructioncolor}{rgb}{.5,.5,.5}

\usepackage[font=small]{caption}

\def\secref#1{Sec.~\ref{#1}}
\def\figref#1{Fig.~\ref{#1}}
\def\tabref#1{Tab.~\ref{#1}}
\def\eqref#1{Eq.~(\ref{#1})}

\makeatletter
\usepackage{xspace}
\DeclareRobustCommand\onedot{\futurelet\@let@token\@onedot}
\def\@onedot{\ifx\@let@token.\else.\null\fi\xspace}
\def\eg{e.g\onedot}

\def\etal{{et al}\onedot}
\makeatother

\def\etalcite#1{\etal~\cite{#1}}

\usepackage{array}
\newcolumntype{L}[1]{>{\raggedright\let\newline\\\arraybackslash\hspace{0pt}}m{#1}}
\newcolumntype{C}[1]{>{\centering\let\newline\\\arraybackslash\hspace{0pt}}m{#1}}
\newcolumntype{R}[1]{>{\raggedleft\let\newline\\\arraybackslash\hspace{0pt}}m{#1}}

\newcommand{\RR}{\mathbb{R}}

\renewcommand{\b}[1]{\mbox{\boldmath$#1$}}

\newcommand{\set}[1]{\mathcal{#1}} 	

\renewcommand{\vec}[1]{{\b #1}}

\newcommand{\m}[1]{{\mbox{{\sffamily\slshape{#1\/}}}}}

\usepackage{subfigure}
\usepackage{multirow}
\usepackage{rotating}  % for rotating sideway

\title{SeqOT: A Spatial-Temporal Transformer Network \\ \mbox{for Place Recognition Using Sequential LiDAR Data}}

\author{Junyi Ma, Xieyuanli Chen, Jingyi Xu, Guangming Xiong$^*$ % <-this % stops a space
  \thanks{
  J. Ma, J. Xu, and G. Xiong are with the Beijing Institute of Technology. X. Chen is with the National University of Defense Technology.}
  \thanks{$^*$corresponding author email: xiongguangming@bit.edu.cn}
}

\begin{document}
\maketitle

\IEEEpeerreviewmaketitle
%\thispagestyle{empty}
%\pagestyle{empty}

%%%%%%%%%%%%%%%%%%%%%%%%%%%%%%%%%%%%%%%%%%%%%%%%%%%%%%%%%%%%%%%%%%%%%%%%%%%%%%%%
\begin{abstract}

Place recognition is an important component for autonomous vehicles to achieve loop closing or global localization. In this paper, we tackle the problem of place recognition based on sequential 3D LiDAR scans obtained by an onboard LiDAR sensor. We propose a transformer-based network named SeqOT to exploit the temporal and spatial information provided by sequential range images generated from the LiDAR data. It uses multi-scale transformers to generate a global descriptor for each sequence of LiDAR range images in an end-to-end fashion. During online operation, our SeqOT finds similar places by matching such descriptors between the current query sequence and those stored in the map. We evaluate our approach on four datasets collected with different types of LiDAR sensors in different environments. The experimental results show that our method outperforms the state-of-the-art LiDAR-based place recognition methods and generalizes well across different environments. Furthermore, our method operates online faster than the frame rate of the sensor. 
The implementation of our method is released as open source at: \url{https://github.com/BIT-MJY/SeqOT}.

\end{abstract}


%%%%%%%%%%%%%%%%%%%%%%%%%%%%%%%%%%%%%%%%%%%%%%%%%%%%%%%%%%%%%%%%%%%%%%%%%%%%%%%%
\section{Introduction}
\label{sec:intro}

Given a map, place recognition can be defined as whether a robot can recognize where the currently observing place is in the map. It plays an important role for most navigation systems during simultaneous localization and mapping (SLAM) or global localization. 
Vision-based place recognition has been well-discussed in existing research papers~\cite{yin2019icra, hausler2021patch, lowry2016tro}. However, cameras are usually influenced by illumination and seasonal changes, thus less reliable when used in large-scale outdoor environments. In contrast, LiDAR-based place recognition is more robust to such changes and attracts many research interests in recent years~\cite{ma2022ral, chen2020rss, kim2018scan}. To further enhance the robustness of long-term place recognition, there are several works~\cite{milford2012icra, liu2019seqlpd, garg2021ral} exploiting sequential sensor data. They generate descriptors for every single observation and mainly focus on matching the sequence of such descriptors. 

In this paper, we propose a novel sequential data-enhanced place recognition method named SeqOT based on our previous work OverlapTransformer (OT) \cite{ma2022ral}. It utilizes sequences of range images generated from 3D LiDARs mounted on autonomous vehicles to achieve online place recognition. Unlike the existing methods using sequence-based descriptors matching, our proposed SeqOT uses multi-scale transformers and generates only a single global descriptor for each sequence of LiDAR range images in an end-to-end fashion. It first uses a devised single-scan transformer module to extract features of each scan. Then, a multi-scan transformer module is applied to fuse spatio-temporal information and generate sub-descriptors for every continuous three scans. In the end, it exploits a pooling module to fuse sub-descriptors into a final descriptor of the LiDAR sequence, which is finally used to find the correct places in the map efficiently and reliably.
%Following our previous work~\cite{chen2020rss,chen2021auro,ma2022ral}, we apply the overlap concept, which represents the similarity of the scan pairs, to supervise the network learning and evaluate place recognition performance.

\begin{figure}
\vspace{0.2cm}
  \centering
  \includegraphics[width=1\linewidth]{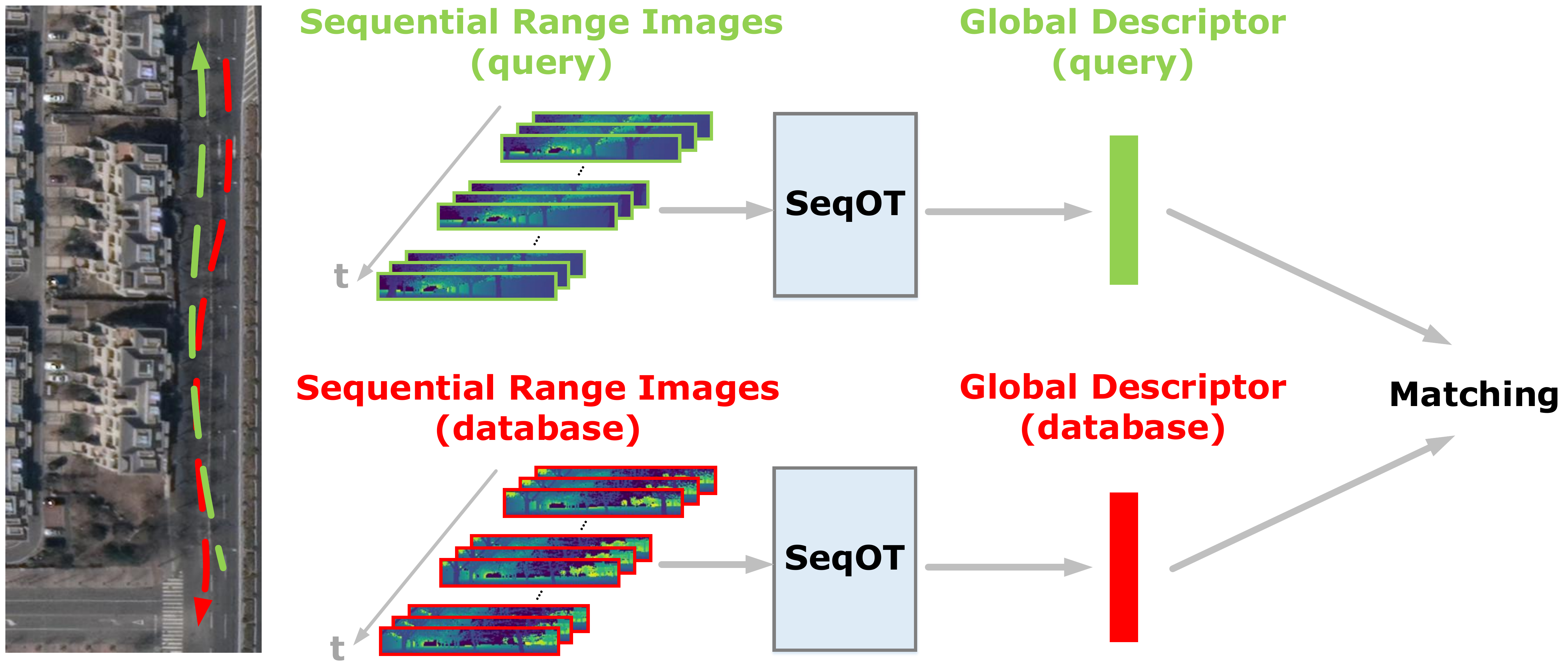}
  \caption{Our proposed SeqOT exploits sequential range images from LiDAR sensors as input to extract spatial and temporal features at the same time, and generate a final sequence enhanced global descriptor.}
  \label{fig:motivation}
  \vspace{-0.6cm}
\end{figure}

The main contribution of this paper is an end-to-end network that exploits sequential LiDAR range images to achieve reliable long-term place recognition performance. It exploits multi-scale transformer modules to fuse spatial and temporal information of sequential LiDAR data and generate global descriptors for fast retrieval. Benefiting from the proposed yaw-rotation-invariant architecture, SeqOT is robust to the viewpoint change and the order of input scans, thus achieving reliable place recognition even when the car drives in opposite directions. 
We thoroughly evaluated our SeqOT on the Haomo and NCLT datasets. The experimental results show the superiority of our method compared with both state-of-the-art single-scan and sequence-enhanced baseline methods. We also provide zero-shot place recognition results on the KITTI and MulRan datasets using the model pre-trained with the NCLT dataset to show the good generalization ability of our method. \change{To thoroughly evaluate our approach, we furthermore provide the ablation studies on the sequence length, the transformer modules, and the yaw-rotation invariance.}

In sum, we make four claims that our approach is able to
(i)~achieve good long-term place recognition in outdoor large-scale environments using only sequential LiDAR data,
(ii)~generalize well into the different environments using LiDAR data obtained from different types of LiDAR sensors without fine-tuning,
(iii)~recognize places with changing viewpoints and input sequence order based on the proposed yaw-rotation-invariant architecture,
(iv)~achieve online operation with runtime less than 100\,ms.

%%%%%%%%%%%%%%%%%%%%%%%%%%%%%%%%%%%%%%%%%%%%%%%%%%%%%%%%%%%%%%%%%%%%%%%%%%%%%%%%
\section{Related Work}
\label{sec:related}

Place recognition is a classic topic in robotics and computer vision~\cite{vysotska2019ral, arandjelovic2016netvlad, hausler2021patch, lowry2016tro, vysotska2017irosws}. In this paper, we focus mainly on LiDAR-based methods and refer vision-based place recognition to the survey paper by Lowry~\etalcite{lowry2016tro}.	
 
Many works have been done in LiDAR-based place recognition and loop closure detection due to their accurate range measurements and the robustness of LiDAR sensors toward illumination changes.
For example, Kim \etal~\cite{kim2018scan} propose Scan Context (SC) to count the maximum height of separated point clouds to generate a 2D image as a descriptor for finding similar scans. Delight by Cop \etal~\cite{cop2018delight} and ISC by Wang \etal~\cite{wang2020intensity} both utilize geometric and remission information of LiDAR data to enhance the discrimination of generated descriptors. Instead of using LiDAR data in spatial domain, LiDAR Iris by Wang \etal~\cite{wang2020lidar} and BVMatch by Luo \etal~\cite{luo2021ral} transform the point clouds into the frequency domain and generate yaw-rotation-invariant descriptors to find scans in the same place taken from different viewpoints. 
There are also works~\cite{chen2021icra,dong2021online} using LiDAR range images to calculate the similarities between LiDAR scans and later combined with Monte Carlo localization to achieve global localization. 

Besides the above-mentioned hand-crafted methods, learning-based methods have recently attracted attention with the advent of neural networks. PointNetVLAD by Uy \etal~\cite{uy2018pointnetvlad} is the first method introducing the NetVLAD~\cite{arandjelovic2016netvlad} architecture into learnable LiDAR-based place recognition. Based on NetVLAD, Liu \etal~\cite{liu2019lpd} propose LPD-Net using multiple local features extracted from a point cloud as input of graph neural network to generate global descriptors. In contrast, OverlapNet by Chen~\etalcite{chen2020rss,chen2021auro} utilizes siamese networks to estimate the similarity between a pair of LiDAR scans, and later uses such similarities for global localization~\cite{chen2020iros}. \change{Kong \etal~\cite{kong2022tie} later also propose a siamese network with multiple cues as input to learn a submap-based model. More recently, Cao \etal~\cite{cao2020tie, cao2022tie} propose cylindrical-image-based methods to mitigate the influence of viewpoint changes.}
%More recently, DiSCO by Xu \etal~\cite{xu2021ral} and SCI proposed by Kim \etal~\cite{kim2019ral} combine deep learning with frequency domain to generate robust descriptors to find similar LiDAR scans. 
There are also works using high-level semantic information to improve LiDAR-based place recognition results, such as SGPR by Kong \etal~\cite{kong2020semantic}, and RINet by Li \etal~\cite{li2022ral_rinet}. However, such high-level semantic information is not always available in different environments.
To exploit point cloud 3D spatial information for place recognition, Minkloc3D by Komorowski~\cite{Komorowski2021wacv} and LoGG3D-Net by Vidanapathirana~\etal~\cite{vid2022icra} use sparse convolution for effectively extracting point-wise features. More recently, attention mechanism~\cite{vaswani2017nips} has been introduced to generate more discriminative descriptors for LiDAR-based place recognition. For example, PPT-Net by Hui \etal~\cite{hui2021iccv}, NDT-Transformer by Zhou \etal~\cite{zhou2021ndt}, and our previous work OverlapTransformer~\cite{ma2022ral} use the transformer network~\cite{vaswani2017nips} to enhance the descriptiveness of descriptors and improve the place recognition performance. 

Single observation-based place recognition methods have achieved reasonable results. However, they are less reliable when used for long-time-span place recognition with appearance changes.
\change{Multiple vision-based place recognition methods use sequential/temporal images to enhance the performance for long-term place recognition.} For example, SeqSLAM by Milford and Wyeth~\cite{milford2012icra} matches sequences of images using the normalized sum of pixel intensity differences. Based on that, Fast-SeqSLAM by Siam \etal~\cite{siam2017icra} uses a nearest neighbor algorithm to find possible candidates, and reduces time complexity significantly for long-term visual place recognition. In contrast, Vysotska \etal~\cite{vysotska2017irosws} propose a hashing-based image retrieval strategy for more efficient sequence-based relocalization. More recently, Garg \etal~\cite{garg2021ral} propose SeqNet, which combines the convolution network with a simplified version of SeqSLAM~\cite{milford2012icra} to better use the relations between sequences of descriptors. They later also propose SeqMatchNet~\cite{garg2021corl}, which uses a triplet loss to supervise a convolutional network for image sequence matching. It improves the optimization process by using a sequence-based distance metric instead of single summary vector metrics compared with previous methods.

Although the sequence-enhanced methods have been well-studied in vision-based place recognition, there are not many works for LiDAR-based methods. 
The early work SeqLPD by Liu \etal~\cite{liu2019seqlpd} exploits the lightweight version of their previous single-scan-based LiDAR place recognition network LPD-Net~\cite{liu2019lpd} together with a coarse-to-fine sequence matching approach to recognize places using LiDAR sequences. More recently, Yin \etal~\cite{yin2021fusionvlad} propose FusionVLAD to generate multi-view representations with dense submaps from sequential LiDAR scans, and encode both the top-down and spherical views of LiDAR scans. They later also propose SeqSphereVLAD~\cite{yin2022tie,yin2020iros}, which locates the best match using a particle filter-based method in the global searching thus improving the place recognition robustness. These methods have achieved comparable performance by combining single-scan/submap-based place recognition methods with sequence matching. \change{However, they exploit the spatial and temporal information of sequential scans in a two-step manner and may not fuse them very well. Besides, they are not robust enough to yaw-rotation in real vehicle applications. In this paper, we propose an end-to-end sequence-enhanced place recognition method. Different to the existing sequence-enhanced methods \cite{vysotska2017irosws, liu2019seqlpd, yin2022tie, yin2020iros, garg2021ral}, we fuse the spatial and temporal information using yaw-rotation-invariant transformer networks, and directly generate one single global descriptor for each LiDAR sequence in an end-to-end fashion for fast LiDAR sequence-based place recognition.}

\begin{figure*}[!t]
  %\vspace{0.2cm}
  \centering
  \includegraphics[width=\linewidth]{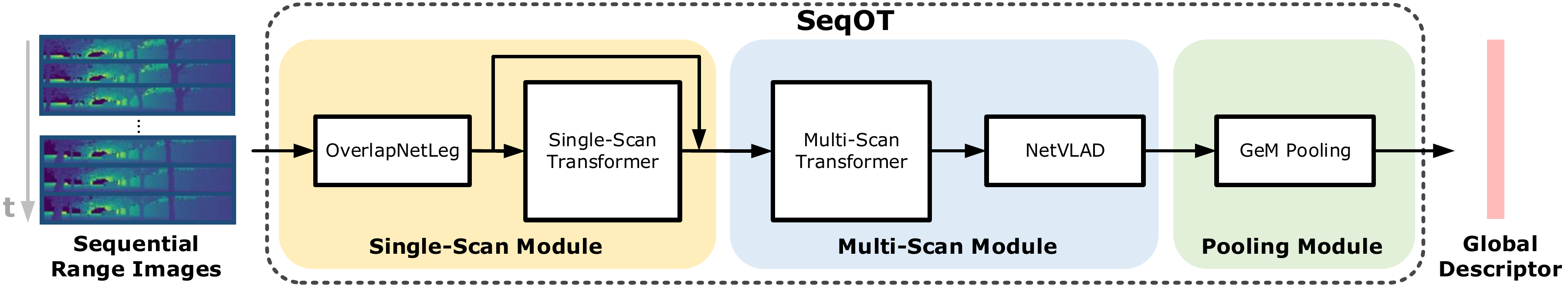}
  \caption{Pipeline overview of our proposed SeqOT. It first uses the single-scan module to extract features for each range image in the sequence Then, the multi-scan module fuses every three continuous features to exploit the spatial-temporal information and generates a sub-descriptor. In the end, a GeM pooling module is applied to fuse the sub-descriptors and generate a global descriptor for fast LiDAR sequence-based place recognition.}
  \label{fig:pipeline}
  \vspace{-0.4cm}
\end{figure*}

%%%%%%%%%%%%%%%%%%%%%%%%%%%%%%%%%%%%%%%%%%%%%%%%%%%%%%%%%%%%%%%%%%%%%%%%%%%%%%%%
\section{Our Approach}
\label{sec:overlap_network}

%%%%%%%%%%%%%%%%%%%%%%%%%%%%%%%%%%%%%%%%%
%\subsection{System Overview}

The overview of our proposed method is depicted in~\figref{fig:pipeline}. Our proposed SeqOT is composed of three modules, including a single-scan module, a multi-scan module, and a pooling module. It takes sequential range images in the queue as the input of the proposed spatial-temporal network and first extracts features from every single LiDAR range image using the single-scan module. Then, it uses the multi-scan module to generate a sub-descriptor of every continuous features of three scans. In the end, a GeM pooling module is applied to fuse the sub-descriptors into a 1-D global descriptor for fast LiDAR-sequence-based place recognition. More details about the architecture design are presented in~\secref{sec:seqot}.
Benefiting from the devised network architecture, we mathematically show in~\secref{sec:ayai} that our proposed descriptor is yaw-rotation-invariant.
For training the proposed network, we propose a two-phase training method and use the triplet loss with the calculated overlap to better distinguish the positive and negative training examples presented in ~\secref{sec:loss}.

%%%%%%%%%%%%%%%%%%%%%%%%%%%%%%%%%%%%%%%%
\subsection{SeqOT Network Architecture}
\label{sec:seqot}

Our method takes only sequential LiDAR data as the input. Following our previous work~\cite{ma2022ral,chen2021auro}, we use the range image representation of LiDAR scans. The correspondence between a LiDAR point $\vec{p}_i \in \set{P}, \vec{p}_i = (x_i, y_i, z_i) $ and the pixel coordinates $(u_i,v_i)$ in the corresponding range image $\set{R}$ can be represented as:
\begin{align}
  \left( \begin{array}{c} u_i \vspace{0.0em}\\ v_i \end{array}\right) & = \left(\begin{array}{cc} \frac{1}{2}\left[1-\arctan(y_i, x_i)/ \pi\right] w   \vspace{0.5em} \\
      \left[1 - \left(\arcsin(z_i/r_i) + \mathrm{f}_{\mathrm{up}}\right)/ \mathrm{f}\right] h\end{array} \right), \label{eq:projection}
\end{align}
where $r_i = ||\vec{p}_i||_2$ is the range measurement of the corresponding point,~$\mathrm{f} = \mathrm{f}_{\mathrm{up}} + \mathrm{f}_{\mathrm{down}}$ is the vertical field-of-view of the sensor, and~$w, h$ are the width and height of the resulting range image. The range image is dense but light-weighted, which enables our method to operate online.

Unlike the previous works~\cite{ma2022ral,chen2021auro} that only use one LiDAR scan as input, our SeqOT uses a sequence of range images to exploit both spatial and temporal information. At the current timestamp $t$, we use $m$ past range images~\mbox{$\RR_t=\{\set{R}_{t-m+1}, ..., \set{R}_{t-1}, \set{R}_t\}$}.
For all the range images in the sequence~$\RR_t$, we first use a single-scan module to extract features that exploit the spatial information contained in the single range image. 
Following OverlapTransformer~\cite{ma2022ral}, we use an encoder called OverlapNetLeg consisting of several fully convolutional layers to compress the range image with the size of $1\times h\times w$ into a feature volume with the size of $c\times 1 \times w$. The extracted feature keeps the same width as the original range image while compressing the height to $1$ and expanding the channel size to $c$. By doing so, we extract a coarse feature of the range image while maintaining the horizontal information to enable yaw-rotation invariance. Then, we apply a transformer module on such feature volume to further exploit the spatial information and generated a refined feature. 
\change{We concatenate it with the coarse feature to exploit information obtained by both the convolutional layers and attention mechanisms and use the combined feature as the input to the following module.}

After generating the features for all range images in the sequence, we apply the multi-scan module on every three scans to exploit the temporal information. To this end, we concatenate features of every three scans along the width yielding a long feature with the size of $2c \times 1 \times 3w$ and use these long features as the input to the multi-scan module. It first applies the self-attention mechanism~\cite{vaswani2017nips} of the multi-scan transformer module, and then a NetVLAD with MLP~\cite{ma2022ral} on the output feature of the transformer yielding a $1\times 256$ sub-descriptor vector which fuses the spatial and temporal information together.

There are two transformer modules utilized separately in the proposed single-scan module and multi-scan module. They have the same architecture but play different roles. Both transformers consist of a multi-head self-attention (MHSA), a feed-forward network (FFN), and layer normalization (LN). 
The input feature of MHSA is split into several sub-features along the channel dimension according to the number of heads. 
The so-called query, key, and value $\{Q, K, V\}$ features of MHSA are assigned as the same as the input feature with multi-head sub-features, and the dot-product attention mechanism of MHSA can be formulated as:
\begin{align}
  \bar{A}&=\text{Attention}(Q, K, V) 
  =\text{softmax} \Big(\frac{Q {K}^{T}}{\sqrt{d_k}}\Big) V \vspace{0.5em} ,
 \label{eq:ScaledDotProductAttention}
\end{align}
where $d_k$ represents the dimension of splits. \change{$\bar{A}$ is the output feature volume of the single-head self-attention in MHSA. The output of MHSA is then fed to the FFN for position-wise linear transformations. The LN for a layer-wise normalization yielding the final output feature $A$ of the transformer.}
We refer more details to the original transformer paper~\cite{vaswani2017nips}.

The single-scan transformer module is designed to exploit the spatial information of a single scan and generate an intermediate feature for each scan. The multi-scan transformer module operates on the concatenated features from consecutive three observations focusing on the relations of temporal features. Different from existing spatial-temporal networks ~\cite{garg2021ral,mersch2021corl,toyungyernsub2021icra} using convolution networks on multiple scans, we use two transformers at different scales to capture both local information of a single scan and sequential relations between multiple scans. 
We apply the multi-scan transformer on three consecutive scans to exploit spatial-temporal information while keeping the network lightweight and efficient. 

In the end, we use a generalized mean (GeM) pooling~\cite{radenovic2018tpami} to fuse all the sub-descriptors generated by the multi-scan module to further exploit the temporal information in an even longer range. The GeM pooling fuses the sub-descriptors into one global descriptor along the time dimension, which is invariant to the input order of the sub-descriptors. Based on such lightweight global descriptors, our method achieves very fast LiDAR-sequence-based descriptor matching.

\subsection{Yaw-Rotation Invariance}
\label{sec:ayai}

Our devised SeqOT is yaw-rotation-invariant, leading to more robust place recognition performance in real applications, \eg, an autonomous vehicle operating on two-way streets. 
In this section, we provide the mathematical derivation to show that our SeqOT is yaw-rotation-invariant against sequential LiDAR scans obtained at the same place but from different viewpoints. 

It has been proved in OverlapTransformer~\cite{ma2022ral} that the single-scan transformer module is yaw-rotation equivariant. The yaw rotation $\theta$ of a point cloud $\set{P}$ leads to the horizontal shift $s$ of the corresponding feature volume $X$. We follow OverlapTransformer~\cite{ma2022ral} to use the term $\m{C}_s$ represents the column shift of the feature volume $X$ by matrix right multiplication, and $\m{R}_\theta$ represents the yaw rotation matrix of the point cloud $\set{P}$ and the following equation holds:
%\begin{align}
%  {X}C_s&=\text{Attention}_\text{SS}(R_\theta\set{P})~,
%\end{align}
\begin{align}
  X\m{C}_s&=\text{SSM}(\m{R}_\theta\set{P})~,
  \label{eq:single}
\end{align}
where $\text{SSM}(\cdot)$ represents the single-scan module operation.

\begin{figure}[t]
  %\vspace{0.2cm}
  \centering
  \includegraphics[width=1\linewidth]{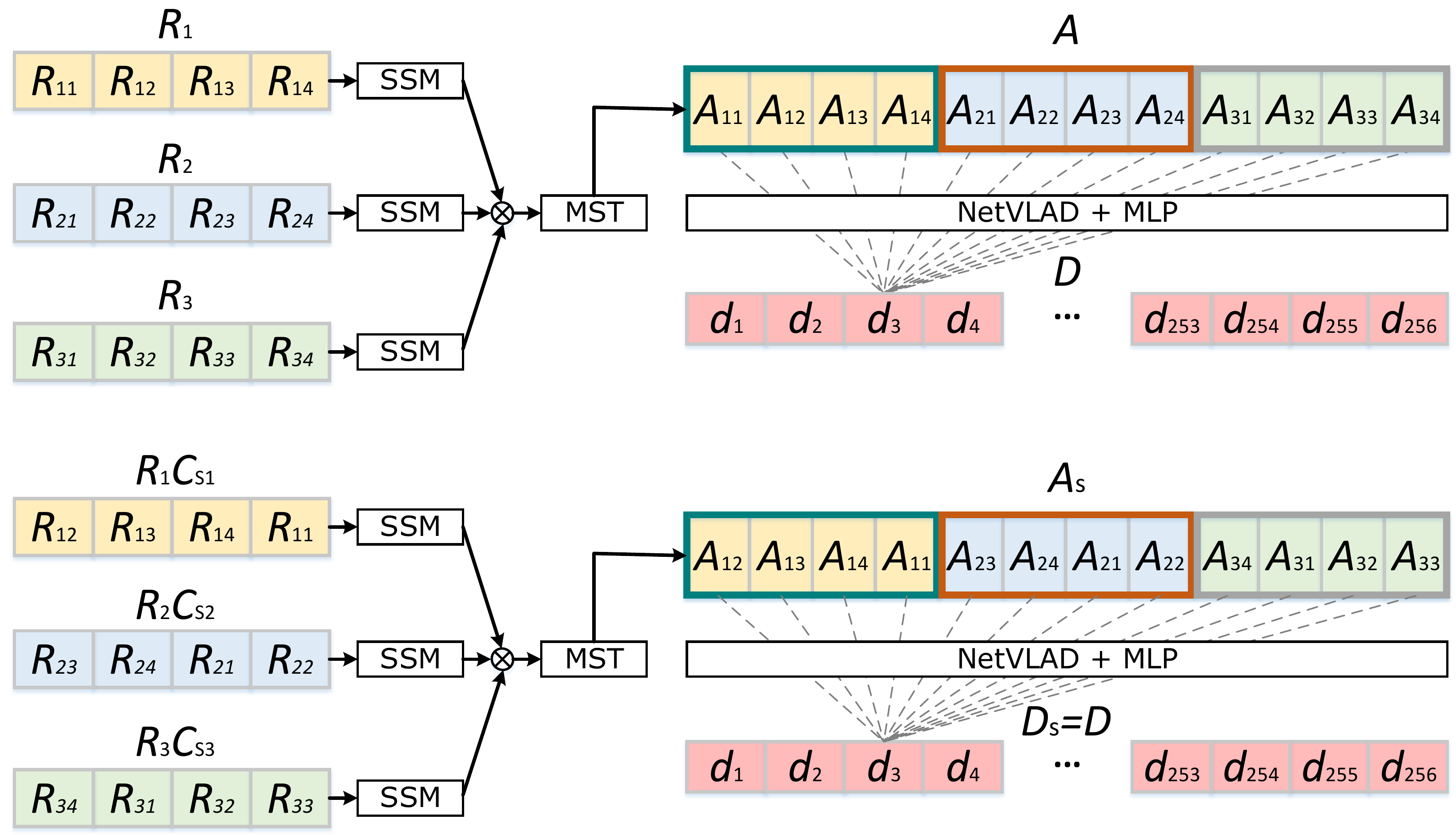}
  \caption{The output of the multi-scan transformer (MST) is yaw-rotation-equivariant individually, leading to the same sub-descriptor generated by the NetVLAD with MLP.}
  \label{fig:partially_yaw_equi}
    \vspace{-0.45cm}
\end{figure}

After obtained the features from the single-scan module, our SeqOT concatenates three feature volumes ${X}_1, {X}_2, {X}_3$ of three consecutive scans along the width and generates a long feature volume ${X}_\text{long} = [{X}_1, {X}_2, {X}_3]$. 
%It then applys the multi-scan transfomer on such long feature volumes.
%The MHSA in the multi-scan transfomer can be fomulated as 
%\begin{align}
%  \set{A}&=\text{Attention}_\text{MS}(Q, K, V) 
%  =\text{softmax} \Big(\frac{Q {K}^{T}}{\sqrt{d_k}}\Big) V \vspace{0.5em} ,
% \label{eq:ScaledDotProductAttention}
%\end{align}
%\begin{align}
%  \set{A}&=\text{Attention}_\text{MSM}(Q, K, V) 
%  =\text{softmax} \Big(\frac{Q {K}^{T}}{\sqrt{d_k}}\Big) V \vspace{0.5em} ,
% \label{eq:ScaledDotProductAttention}
%\end{align}
%where $\{Q, K, V\}$ are the query, key and value feature splits along the channels of the input feature volume ${X}_\text{long}$. $d_k$ represents the dimension of splits. $\set{A}$ is the output feature volume of the multi-scan transfomer.
According to~\eqref{eq:single}, the yaw-rotations of the raw input point clouds lead to shifted feature volumes $[{X}_1C_{s1}, {X}_2C_{s2}, {X}_3C_{s3}]$. The feature volume $\bar{A}_s$ extracted by the MHSA of the multi-scan transformer in~\eqref{eq:ScaledDotProductAttention} then becomes:
%\begin{footnotesize}
\begin{align}
  \bar{A}_s &= \text{Attention}(Q_s, K_s, V_s) = \text{softmax} \Big(\frac{Q_s {K_s}^{T}}{\sqrt{d_k}}\Big) V_s \\
  &= \text{softmax} 
  \Bigg(
  \frac{[{X}_1C_{s1}, {X}_2C_{s2}, {X}_3C_{s3}] 
  \begin{bmatrix}
  	C_{s1}^{T}{X}_1^{T} \\ 
  	C_{s2}^{T}{X}_2^{T} \\ 
  	C_{s3}^{T}X_3^{T} 
  \end{bmatrix}}
  {\sqrt{d_k}}
  \Bigg)  \nonumber \\ 
  & \quad\quad [{X}_1C_{s1}, {X}_2C_{s2}, {X}_3C_{s3}] \\
  &= \text{softmax} \Bigg(\frac{{X}_1{X}_1^{T}+{X}_2{X}_2^{T}+{X}_3{X}_3^{T}}{\sqrt{d_k}}\Bigg) \nonumber \\ 
  & \quad\quad [{X}_1C_{s1}, {X}_2C_{s2}, {X}_3C_{s3}] \\
  &= W[{X}_1C_{s1}, {X}_2C_{s2}, {X}_3C_{s3}] \\
  &= [W{X}_1C_{s1}, W{X}_2C_{s2}, W{X}_3C_{s3}] \,,
 \label{eq:ShiftedScaledDotProductAttention}
\end{align}
%\end{footnotesize}
where $W$ is a weighting matrix generated by the softmax and not affected by feature shifting $C_s$. As can be seen, the new feature $\bar{A}_s$ also consists of three sub-features \mbox{$\bar{A}_s = [W{X}_1C_{s1}, W{X}_2C_{s2}, W{X}_3C_{s3}]$}, where each sub-feature is yaw-rotation-equivariant individually. The LN and FFN also does not affect the yaw-rotation equivariance of each part~\cite{ma2022ral}, and we have $A_s = f(\bar{A}_s)$ where $f(\cdot)$ represents the following operations after MHSA in the transformer module.
Therefore, the yaw rotation in point clouds only causes the permutation change along the width dimension in the output feature of the multi-scan transformer.
\change{Since the following NetVLAD module is permutation invariant~\cite{ma2022ral, uy2018pointnetvlad}, it generates the yaw-rotation-invariant sub-descriptor $D$ as:}
\begin{align}
  D_s &= \text{Vlad}(A_s) \\
  &= \text{Vlad}(f([W{X}_1C_{s1}, W{X}_2C_{s2}, W{X}_3C_{s3}])) \\ 
  &= \text{Vlad}(f([W{X}_1, W{X}_2, W{X}_3])) \\
  &= \text{Vlad}(A) = D\,,
 \label{eq:netvlad}
\end{align}
which is also illustrated in~\figref{fig:partially_yaw_equi}. 

\subsection{Network Training}
\label{sec:loss}

One of the bottlenecks of using neural networks on sequential LiDAR data is the large computational and memory costs. To tackle this problem, we use a two-phase scheme to train the transformer modules and the pooling module separately. \change{For a recorded LiDAR range image sequence $\mathbb{R}_\text{all} = \{\set{R}_i\}^n_{i=0}$ with $n+1$ scans, we group each scan $\set{R}_i$ with its $l-1$ adjacent scans to form a sub-sequence training sample, where $l=3$ is the same number as the input scans of our multi-scan module. For each training sample, we use the overlap/similarity value proposed by~\cite{chen2021auro} between the anchor scan $\set{R}_i$ and all other anchor scans in $\mathbb{R}_\text{all}$ to supervise the network.} The overlap value between two anchor scans $\set{R}_i$ and $\set{R}_j$ is calculated as:
\begin{align}
  O_{\set{R}_i \set{R}_j} & = \frac{\sum_{(u,v)} \mathbb{I}\Big\{\left|\left| \set{R}_i(u,v) - \set{R}^{\prime}_j(u,v) \right|\right| \leq \delta \Big\}} {\min \left( \textrm{valid}(\set{R}_i), \textrm{valid}(\set{R}^{\prime}_j) \right)},  \label{eq:overlap}
\end{align}
\change{where $\set{R}^{\prime}_j$ is the reprojected range image of $\set{R}_j$ in the coordinate frame of the query scan $\set{R}_q$.} $\mathbb{I}(a)=1$ if $a$ is true and $\mathbb{I}(a)=0$ otherwise, $\text{valid}(\set{R})$ refers to the counts of valid pixels of range image $\set{R}$, and  $\delta$ is the threshold to decide the overlapped pixel.

We train the transformer networks using the triplet loss. If the overlap value between two training samples is larger than 0.3, we take that pair as a positive sample, otherwise a negative sample. 
 
In the first phase, we train the network without the pooling module. The triplet loss is directly applied on the sub-descriptors generated by our multi-scan module. For each query sub-descriptor $D_\text{q}$, $N_\text{pos}$ positive sub-descriptors \mbox{$\set{D}_\text{pos} = \{D_{p}\}$}, and $N_\text{neg}$ negative sub-descriptors \mbox{$\set{D}_\text{neg} = \{D_n\}$} are used to calculate the triplet loss by:
\begin{align}
& \set{L}_1(D_\text{q},\set{D}_\text{pos},\set{D}_\text{neg})=    \nonumber \\
& N_\text{pos}(\alpha+\max_{p}(d(D_\text{q},D_p))) 
- \sum_{N_\text{neg}}(d(D_\text{q},D_n)),
\label{eq:tripletloss}
\end{align}
where $\alpha$ is the margin to keep the loss positive and $d(\cdot)$ computes the squared Euclidean distance.
We use the triplet loss to minimize the distance between the query sub-descriptor and the hardest case in positive reference sub-descriptors, and maximize the distance between the query and all sampled negative reference sub-descriptors.

In the second phase, we use the pre-trained transformer network to generate sub-descriptors $\set{D}_\text{all}$ for all the scans in $\mathbb{R}_\text{all}$. We then train the GeM Pooling with the sub-descriptors $\set{D}_\text{all}$ using the same overlap labels used in the first phase. For each query global descriptor $G_\text{q}$, $N_\text{pos}$ positive descriptors \mbox{$\set{G}_\text{pos} = \{G_p\}$}, and $N_\text{neg}$ negative descriptors \mbox{$\set{G}_\text{neg} = \{G_n\}$} are used to calculate the triplet loss by:
\begin{align}
& \set{L}_2(G_\text{q},\set{G}_\text{pos},\set{G}_\text{neg})=    \nonumber \\
& N_\text{pos}(\alpha+\max_{p}(d(G_\text{q},G_p))) 
- \sum_{N_\text{neg}}(d(G_\text{q},G_n)).
\label{eq:tripletloss}
\end{align}

For online operation, we store the sub-descriptors of past scans generated by our transformer network and apply the pooling module directly on them. Thus only one feedforward is needed for every incoming scan. 

%%%%%%%%%%%%%%%%%%%%%%%%%%%%%%%%%%%%%%%%

\section{Experimental Evaluation}
\label{sec:exp}
The experimental evaluation is designed to showcase the performance of our approach and to support the claims that our approach is able to:
(i)~achieve good long-term place recognition in outdoor large-scale environments using only sequential LiDAR data without any other information,
(ii)~generalize well into the different environments using LiDAR data obtained by different types of LiDAR sensors without fine-tuning,
(iii)~recognize places with changing viewpoints and input sequence order based on the yaw-rotation-invariant architecture,
(iv)~achieve online operation with runtime less than 100\,ms.

\begin{figure}[t]
  %\vspace{0.2cm}
  \centering
  \includegraphics[width=0.85\linewidth]{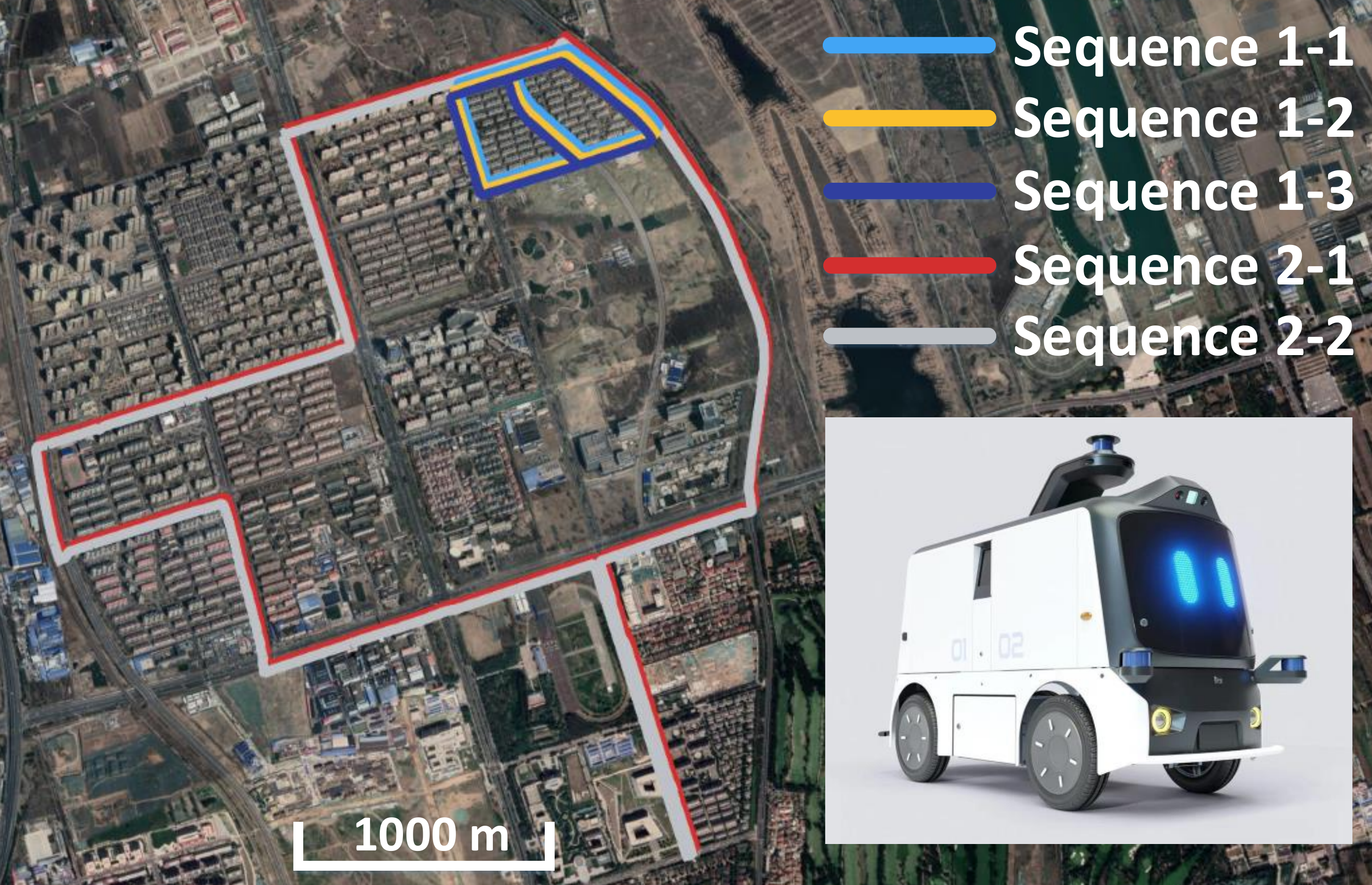}
  \caption{Haomo dataset is collected by an AGV built by HAOMO.AI Technology company in the urban environments of Beijing, which is introduced in detail in our recent work \cite{ma2022ral}.}
  \label{fig:car_cut.jpg}
  \vspace{-0.4cm}
\end{figure}

\subsection{Implementation and Experimental Setup}
We use several different datasets to evaluate our method, including the NCLT~\cite{carlevaris2016ijrr}, KITTI odometry~\cite{geiger2012cvpr}, and MulRan~\cite{kim2020icra} datasets. In addition, we also use our recorded Haomo dataset which was firstly open-source in our recent work \cite{ma2022ral}. Haomo dataset is collected by an AGV equipped with a LiDAR sensor (HESAI PandarXT, 32-beam), a wide-angle camera (SENSING SG2, HFOV $106^{\circ}$, VFOV $56^{\circ}$), an RTK GNSS (Asensing INS570D, $0.1^{\circ}$ error in roll/pitch, $0.2^{\circ}$ error in yaw, $2$ cm $+$ $1$ ppm in RTK position), a mini computer (Intel Xeon E, $3.5$ GHz, $80$ W), and a Nivida Tesla T4 (16-GB memory, $70$ W), which is shown in \figref{fig:car_cut.jpg}.

Following previous works~\cite{chen2021auro,ma2022ral}, we use range images with size of~$1\times32\times 900$. The length of used LiDAR sequence is set to~$m = 20$.
The sub-descriptors generated by our multi-scan module are 1-D vectors with the size of $1\times 256$, and so are the final global descriptors generated by our GeM pooling module.
For training, we set $N_\text{pos}=6$, $N_\text{neg}=6$, and $\alpha=0.5$ for the triplet loss. \change{We train our proposed transformer network without the pooling module for 20 epochs, using the ADAM optimizer to update the weights of the network with an initial learning rate of 5e-6 and a weight decay of 0.9 applied every 5 steps. To train the GeM pooling module, we use the similar configuration but increase the initial learning rate to 5e-5.}

In the following experiments, we compare our proposed SeqOT with both state-of-the-art single-scan-based methods PointNetVLAD~\cite{uy2018pointnetvlad}, MinkLoc3D~\cite{Komorowski2021wacv}, and OverlapTransformer~\cite{ma2022ral}, and sequence-enhanced methods using the features generated by OverlapTransformer, including a simple baseline by averaging single-scan matching scores of sequences (OT\_AVG), the hashing-based retrieval strategy (HRS)~\cite{vysotska2017irosws}, the group-based condition-invariant multi-view (CIMV) method~\cite{facil2019arxiv}, and SeqNet~\cite{garg2021ral}. \change{OT\_AVG is a simplified version of SeqSLAM~\cite{milford2012icra} which only considers the diagonal line of the difference matrix for faster retrieval. It is natural and straightforward to try out the visual-descriptor-based methods directly on LiDAR range images since there are few LiDAR-based sequence-enhanced methods. We, therefore, use CIMV and SeqNet with LiDAR range image descriptors as the baselines and show that our devised novel method works better than directly applying existing methods on LiDAR range images.} We also compare our method with SeqLPD~\cite{liu2019seqlpd} using their open-source implementation. 
The descriptor generated by the original OverlapTransformer is with the size of $1 \times 256$. The output of SeqNet is also set to $1\times 256$, and we set the convolutional kernel tensor of size $256\times 3\times 256$ for the local temporal window of SeqNet for a fair comparison. 
If not explicitly stated otherwise, we use $m = 20$ consecutive scans for all methods in the following experiments. 

Note that, since all sequence-enhanced baseline methods are not order and yaw-rotation-invariant, we apply a forward matching and a reverse matching and choose the better one for comparison. This improves the performance of baseline methods, especially in the case where the car drives in the opposite direction with respect to the database. While our method only matches once, benefiting from our yaw-rotation invariance design and the use of the GeM pooling.

%%%%%%%%%%%%%%%%%%%%%%%%%%%%%%%%%%%%%%%%
\subsection{Evaluation for Place Recognition}
\label{epr}

\begin{table*}[t]
\center
\setlength{\tabcolsep}{8.0pt}
\renewcommand\arraystretch{1.3}
\caption{Comparison of place recognition performance on the three challenges of the Haomo dataset}
\small{
\begin{tabular}{lccccccccc}
\toprule
\multirow{2}{*}{Approach} & \multicolumn{3}{c}{Sequence 1-2}               & \multicolumn{3}{c}{Sequence 1-3}                     & \multicolumn{3}{c}{Sequence 2-1}                     \\ \cline{2-10} 
                          & AR@1        & AR@5        & AR@20 & Recall@1        & AR@5        & AR@20       & AR@1        & AR@5        & AR@20       \\ \hline
PointNetVLAD~\cite{uy2018pointnetvlad}              & 0.913          & 0.936          & 0.957    & 0.400          & 0.521          & 0.623          & 0.861          & 0.932          & 0.944          \\ \hline
%PNV\_AVG                 & 0.978          & 0.985          & 0.989    & 0.454          & 0.493          & 0.581          & 0.878          & 0.940          & 0.952          \\ \hline
%PNV\_HRS~\cite{vysotska2017irosws}                 & 0.934          & 0.981          & 0.986    & 0.417          & 0.498          & 0.635          & 0.883          & 0.933          & 0.951          \\ \hline
%PNV\_CIMV~\cite{facil2019arxiv}                 & 0.951          & 0.966          & 0.984    & 0.435          & 0.548          & 0.702          & 0.892          & 0.942          & 0.953          \\ \hline
%PNV\_SeqNet~\cite{garg2021ral}               & 0.985          & 0.990          & 0.996    & 0.428          & 0.507          & 0.689          & 0.890          & 0.941          & 0.952          \\ \hline
MinkLoc3D~\cite{Komorowski2021wacv}        & 0.948          & 0.971          & 0.993    & 0.450          & 0.653          & 0.808          & 0.858          & 0.924          & 0.940          \\ \hline
OverlapTransformer~\cite{ma2022ral}        & 0.974          & 0.987          & 0.995    & 0.776          & 0.878          & 0.931          & 0.903          & 0.933          & 0.945          \\ \hline
\hline
OT\_AVG~\cite{milford2012icra}                 & 0.985          & 0.989          & 0.998    & 0.803          & 0.843          & 0.895          & 0.894          & 0.943          & 0.964          \\ \hline
HRS~\cite{vysotska2017irosws}                 & 0.980          & 0.988          & 0.998    & 0.806          & 0.877          & 0.949          & 0.901          & 0.941          & 0.961          \\ \hline
CIMV~\cite{facil2019arxiv}                  & 0.990          & 0.999          & \textbf{1.}    & 0.810          & 0.887          & 0.963          & 0.910          & 0.948          & 0.965          \\ \hline
SeqNet~\cite{garg2021ral}                & 0.988          & 0.997          & \textbf{1.}    & 0.816          & 0.879          & 0.962          & 0.906          & 0.946          & 0.963          \\ \hline
SeqLPD~\cite{liu2019seqlpd}                    & 0.982          & 0.996          & \textbf{1.}    & 0.645          & 0.713          & 0.878          & 0.918          & 0.946          & \textbf{0.971} \\ \hline
SeqOT (ours)              & \textbf{0.998} & \textbf{1.} & \textbf{1.}    & \textbf{0.824} & \textbf{0.899} & \textbf{0.967} & \textbf{0.932} & \textbf{0.952} & 0.966          \\ \bottomrule
\end{tabular}}
  \label{tab:pr_on_haomo}
%  \vspace{-0.5cm}
\end{table*}

\begin{table*}[h]
\setlength{\tabcolsep}{3.3pt}
\center
\renewcommand\arraystretch{1.3}
\caption{Comparison of place recognition performance on the NCLT dataset}
\small{
\begin{tabular}{lcccccccccccc}
\toprule
\multirow{2}{*}{Approach} & \multicolumn{3}{c}{2012-02-05}                      & \multicolumn{3}{c}{2012-06-15}                      & \multicolumn{3}{c}{2013-02-23}                      & \multicolumn{3}{c}{2013-04-05}                      \\ \cline{2-13} 
                          & AR@1        & AR@5        & AR@20       & AR@1        & AR@5        & AR@20       & AR@1        & AR@5        & AR@20       & AR@1        & AR@5        & AR@20       \\ \hline
PointNetVLAD~\cite{uy2018pointnetvlad}              & 0.746          & 0.823          & 0.875          & 0.612          & 0.720          & 0.782          & 0.469          & 0.604          & 0.719          & 0.449          & 0.576          & 0.683          \\ \hline
%PNV\_AVG                 & 0.741          & 0.775          & 0.823          & 0.728          & 0.778          & 0.828          & 0.528          & 0.611          & 0.714          & 0.498          & 0.574          & 0.675          \\ \hline
%PNV\_HRS~\cite{vysotska2017irosws}                 & 0.743          & 0.764          & 0.821          & 0.643          & 0.726          & 0.773          & 0.471          & 0.562          & 0.667          & 0.449          & 0.576          & 0.683          \\ \hline
%PNV\_CIMV~\cite{facil2019arxiv}                 & 0.752          & 0.780          & 0.822          & 0.661          & 0.728          & 0.771          & 0.477          & 0.562          & 0.659          & 0.453          & 0.535          & 0.642          \\ \hline
%PNV\_SeqNet~\cite{garg2021ral}                & 0.746          & 0.780          & 0.826          & 0.675          & 0.734          & 0.784          & 0.489          & 0.568          & 0.654          & 0.478          & 0.547          & 0.645          \\ \hline
MinkLoc3D~\cite{Komorowski2021wacv}        & 0.802          & 0.864          & 0.926          & 0.630          & 0.685          & 0.774          & 0.507          & 0.616          & 0.751          & 0.482          &0.587          & 0.685          \\ \hline
OverlapTransformer~\cite{ma2022ral}        & 0.861          & 0.899          & 0.930          & 0.639          & 0.697          & 0.780          & 0.536          & 0.645          & 0.764          & 0.496          & 0.603          & 0.715          \\ \hline
\hline
OT\_AVG~\cite{milford2012icra}                 & 0.876          & 0.925          & 0.952          & 0.610          & 0.718          & 0.840          & 0.561          & 0.698          & 0.830          & 0.529          & 0.658          & 0.791          \\ \hline
HRS~\cite{vysotska2017irosws}                 & 0.869          & 0.925          & 0.955          & 0.624          & 0.716          & 0.821          & 0.557          & 0.669          & 0.814          & 0.498          & 0.643          & 0.752          \\ \hline
CIMV~\cite{facil2019arxiv}                  & 0.871          & 0.925          & 0.957          & 0.642          & 0.730          & 0.852          & 0.564          & 0.707          & 0.835          & 0.527          & 0.663          & 0.797          \\ \hline
SeqNet~\cite{garg2021ral}                & 0.889          & 0.933          & 0.960          & 0.645          & 0.745          & 0.859          & 0.569          & \textbf{0.725} & 0.847          & 0.517          & 0.674          & 0.801          \\ \hline
SeqLPD~\cite{liu2019seqlpd}                    & 0.873          & 0.928          & 0.952          & 0.663          & 0.791          & 0.884          & 0.658          & 0.713          & 0.847          & 0.582          & 0.719          & \textbf{0.835} \\ \hline
SeqOT (ours)              & \textbf{0.917} & \textbf{0.947} & \textbf{0.968} & \textbf{0.762} & \textbf{0.844} & \textbf{0.899} & \textbf{0.691} & 0.723          & \textbf{0.874} & \textbf{0.639} & \textbf{0.724} & 0.826          \\ \bottomrule
\end{tabular}}
  \label{tab:pr_on_nclt}
  \vspace{-0.2cm}
\end{table*}

The first experiment supports our claim that our approach achieves good place recognition in large-scale outdoor environments with long time spans using only sequential LiDAR data without any other information.

Following the experimental setup of OverlapTransformer~\cite{ma2022ral}, we train our approach and baseline methods on the relatively old sequences such as sequences 1-1 and 2-1 in Haomo dataset~\cite{ma2022ral} and sequence 2012-01-08 in NCLT dataset~\cite{carlevaris2016ijrr}, and evaluate on newly records such as sequences 1-2, 1-3 and 2-2 in Haomo dataset, and sequences 2012-02-05,  2012-06-15, 2013-02-23, and 2013-04-05 in NCLT dataset. 
%Following the experimental setup of OverlapTransformer~\cite{ma2022ral}, we train our approach and baseline methods on the relatively old sequences such as sequence 2012-01-08 in NCLT dataset~\cite{carlevaris2016ijrr}, and evaluate on newly records such as sequences 2012-02-05,  2012-06-15, 2013-02-23, and 2013-04-05 in NCLT dataset. In addition to NCLT dataset, we also use Haomo dataset which was firstly open-source in our recent work \cite{ma2022ral}. Haomo dataset is collected by an AGV equipped with a LiDAR sensor (HESAI PandarXT, 32-beam), a wide-angle camera (SENSING-SG2, HFOV $106^{\circ}$, VFOV $56^{\circ}$), and an RTK GNSS in urban environments of Beijing, which is shown in \figref{fig:car_cut.jpg}. Sequences 1-1 and 2-1 in Haomo dataset are used as database, and sequences 1-2, 1-3 and 2-2 are used as query.
Haomo dataset is recorded in a relatively short period of around three months but provides multiple challenges including opposite driving directions, while NCLT dataset provides multiple sequences recorded in the same environment across more than one year and is usually used for evaluating long-term place recognition.

We use average top 1 recall (AR@1), top 5 recall (AR@5), and top 20 recall (AR@20) as the evaluation metrics. \change{For each scan sampled from query sequences, we acquire its top $N$ candidates and ground truth reference scans with overlap values larger than $0.3$ in the database. Once one of the true references is found, we consider that query as a successful loop closure. We can get the recall rate by calculating the proportion of successful retrievals for each sequence.}
The evaluation results are shown in \tabref{tab:pr_on_haomo} and \tabref{tab:pr_on_nclt} for Haomo dataset and NCLT dataset respectively. 
As can be seen, sequence-enhanced methods are generally more robust than single-scan-based methods, and our method outperforms all baseline methods in terms of AR@1 on both datasets and has competitive performance when also considering more place candidates. 
Our SeqOT directly extracts and fuses spatial-temporal information from the sequential LiDAR data which outperforms other sequence-based baselines in most cases in terms of LiDAR sequence-based place recognition. 
Since OverlapTransformer and our SeqOT are both yaw-rotation-invariant, they outperform other counterparts in the opposite driving challenges of sequence 1-3 of the Haomo dataset.
In addition, the experimental results also show that our method works for long time span place recognition. The time gaps between sequences in the NCLT dataset are more than one year, leading to significant appearance changes in the environment. Our proposed SeqOT is designed to capture the temporal and spatial changes of consecutive observations thus not overfitting on specific appearance features. In contrast, OverlapTransformer may focus more on specific spatial features, e.g., a temporarily parked car, without exploiting temporal information thus less reliable in long-term place recognition.

\begin{figure}[t]
  \vspace{0.2cm}
  \centering
  \subfigure[OT]{\includegraphics[width=0.31\linewidth]{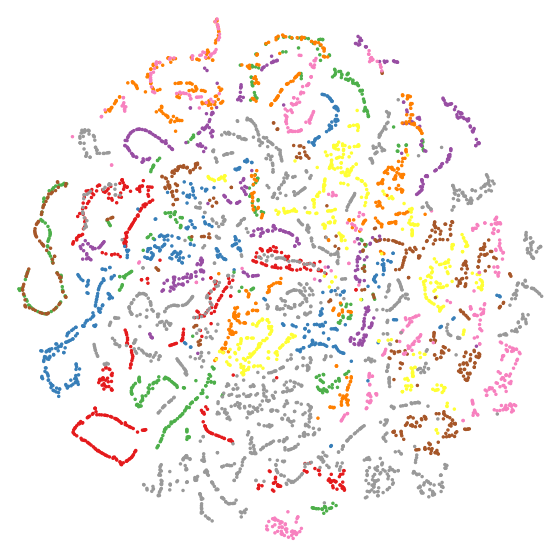}
  \label{fig:tsne_ot}}
  \subfigure[Output of MSM]{\includegraphics[width=0.31\linewidth]{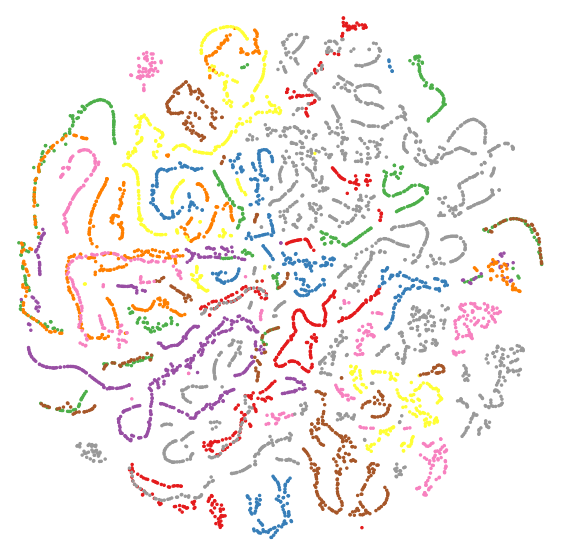}
  \label{fig:tsne_msm}}
  \subfigure[SeqOT]{\includegraphics[width=0.31\linewidth]{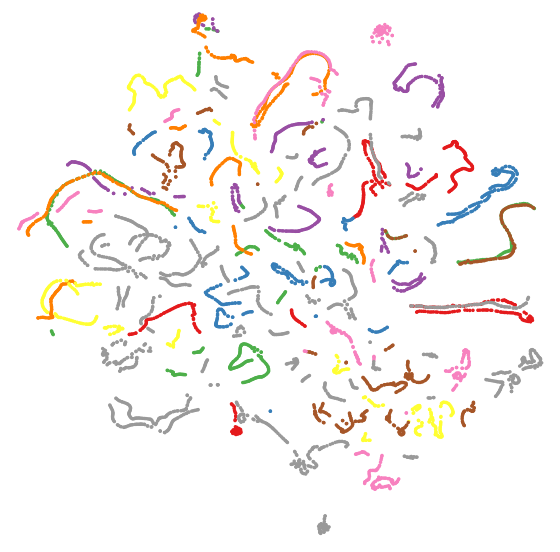}
  \label{fig:tsne_seqot}}
  \vspace{0cm}
  \caption{\change{The t-SNE visualization of place clustering.}}
  \vspace{-0.2cm}
  \label{fig:tsne}
\end{figure}

\change{In addition, we utilize the t-SNE visualization of the global descriptors generated by OT, the output of the multi-scan module (MSM), and the global descriptors generated by SeqOT on sequence 2012-01-08 of the NCLT dataset for qualitative evaluation. As \figref{fig:tsne_ot} and \figref{fig:tsne_msm} show, compared with OT, the t-SNE results of MSM show more distinguishable clusters, which indicates a better feature extraction benefiting from applying the multi-scan transformer. By comparing \figref{fig:tsne_msm} and \figref{fig:tsne_seqot}, we see that with the pooling-based aggregation of sub-descriptors generated by MSM, SeqOT can further generate more discriminative descriptors.}

%%%%%%%%%%%%%%%%%%%%%%%%%%%%%%%%%%%%%%%%
\subsection{Generalization Analysis}

%\begin{figure}[t]
%  %\vspace{0.2cm}
%  \centering
%  \includegraphics[width=0.8\linewidth]{pics_seqot/pr_kitti_new.pdf}
%  \caption{Generalization analysis on the KITTI dataset.}
%  \label{fig:general_anaysis}
%  \vspace{-0.4cm}
%\end{figure}
%\begin{figure}[t]
%  %\vspace{0.2cm}
%  \centering
%  \includegraphics[width=0.8\linewidth]{pics_seqot/pr_mulran_new.pdf}
%  \caption{Generalization analysis on the MulRan dataset.}
%  \label{fig:general_anaysis2}
%  \vspace{-0.4cm}
%\end{figure}

%
%\begin{figure}[t]
%  \vspace{0.2cm}
%  \centering
%  \subfigure[Evaluation on KITTI dataset.]{\includegraphics[width=0.45\linewidth]{pics_seqot/pr_kitti_new.pdf}
%  \label{fig:general_anaysis}}
%  \subfigure[Evaluation on MulRan dataset.]{\includegraphics[width=0.45\linewidth]{pics_seqot/pr_mulran_new.pdf}
%  \label{fig:general_anaysis2}}
%  \caption{Generalization analysis.}
%  \vspace{-0.6cm}
%  \label{fig:overlap_eq3}
%\end{figure}

\begin{figure}[t]
  \vspace{0.5cm}
  \centering
  \subfigure[Evaluation on KITTI dataset.]{\includegraphics[width=0.79\linewidth]{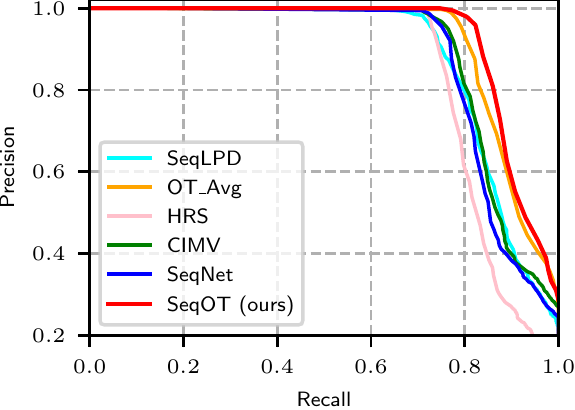}
  \label{fig:general_anaysis}}
  \subfigure[Evaluation on MulRan dataset.]{\includegraphics[width=0.79\linewidth]{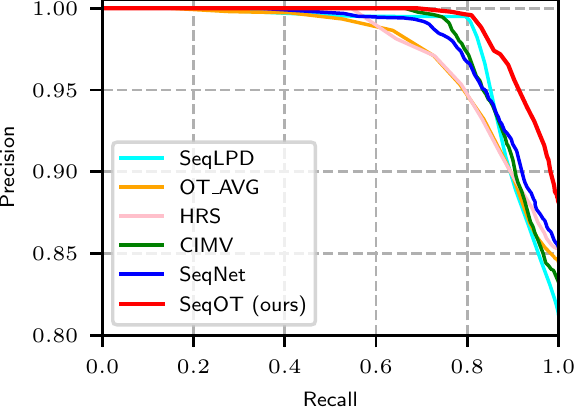}
  \label{fig:general_anaysis2}}
  \caption{\change{Generalization analysis.}}
  \vspace{-0.6cm}
  \label{fig:overlap_eq3}
\end{figure}

In this experiment, we show that our proposed method can generalize well to the different environments with different LiDAR sensors in a zero-shot manner. we train our SeqOT and all the baseline methods only on sequence 2012-01-08 of the NCLT dataset, and operate them directly on the KITTI odometry and MulRan datasets.

We run all methods to find similar places on the areas traveled multiple times in KITTI 00 sequence, while for the MulRan dataset we use sequence kaist-01 as the database and sequence kaist-02 as the query.
Note that the data of the NCLT dataset for training the network was collected using a Velodyne-32 LiDAR, while the KITTI odometry data was collected with Velodyne-64 and the MulRan dataset was with Ouster-64.
In line with OverlapTransformer~\cite{ma2022ral}, we use the precision-recall curve to evaluate all the methods and the experimental results are shown in \figref{fig:general_anaysis} and \figref{fig:general_anaysis2}. As can be seen, SeqOT outperforms other baselines and achieves the best generalization capability. 
The results show the superiority of our SeqOT in both datasets without fine-tuning, which indicates a solid generalization ability.

%%%%%%%%%%%%%%%%%%%%%%%%%%%%%%%%%%%%%%%%
\subsection{Ablation Study on Sequence Length}
\label{sec:ablation_on_seq}

\begin{figure}[t]
  %vspace{0.2cm}
  \centering
  \includegraphics[width=0.79\linewidth]{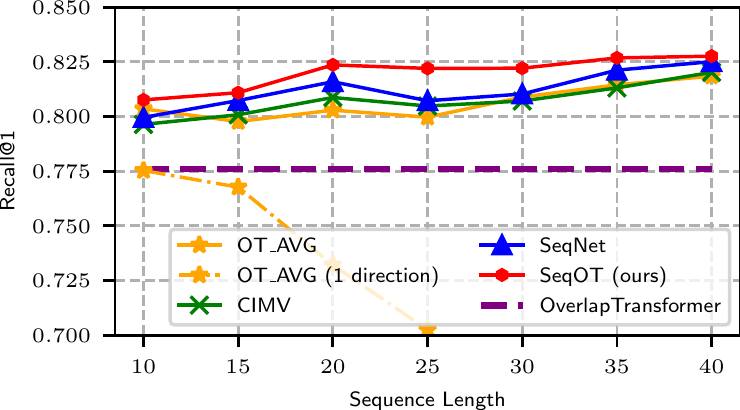}
  \caption{Ablation study on sequence length on the Haomo dataset.}
  \label{fig:aba_on_len_on_haomo}
  \vspace{-0.4cm}
\end{figure}

\begin{figure}[t]
  %\vspace{0.2cm}
  \centering
  \includegraphics[width=0.79\linewidth]{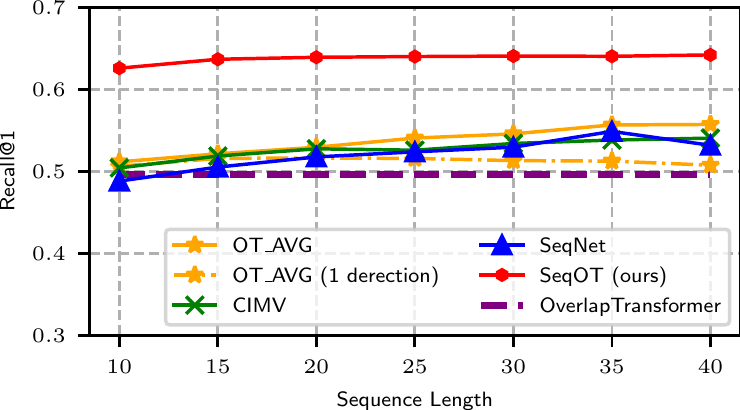}
  \caption{Ablation study on sequence length on the NCLT dataset.}
  \label{fig:aba_on_len_on_nclt}
  \vspace{-0.4cm}
\end{figure}

This ablation study investigates the effect of different lengths of input sequences. 
We change the input sequence length from 10 to 40 LiDAR frames, using OverlapTransformer as the backbone for all the baseline methods. The performance of the single-scan-based OverlapTransformer is also shown as a baseline. We evaluate the average recall AR@1 performance on sequence 1-3 of the Haomo dataset and sequence 2013-04-05 of the NCLT dataset. The results are shown in \figref{fig:aba_on_len_on_haomo} and \figref{fig:aba_on_len_on_nclt}. As can be seen, our proposed SeqOT consistently outperforms all the baseline methods with all tested input lengths. Unlike the baseline methods, our proposed method only applies one forward matching, while the other sequence-enhanced methods need two-direction matching to obtain good performance since the car drives in opposite directions. We also evaluate OT\_AVG with only one forward matching process to show the improvement of two directions matching for baseline sequence matching methods. The experimental results demonstrate that our method is more robust to driving directions benefiting from our devised yaw-rotation-invariant network, while OT\_AVG with one direction matching loses efficacy quickly, especially on the Haomo dataset where the car drives in the opposite directions.
The performance of all methods improves slowly when the sequence length is larger than 20, and the reason could be the redundant information. Therefore, we set the input sequence length to 20 for all the methods in other experiments in trade-off efficiency and performance.

%%%%%%%%%%%%%%%%%%%%%%%%%%%%%%%%%%%%%%%%
\vspace{-0.4cm}
\change{\subsection{Ablation Study on Transformer Modules}}
\label{sec:ablation_on_trans}

\change{This ablation study is aimed to show the effectiveness of the proposed single-scan transformer (SST) and the multi-scan transformer (MST). 
We use sequence 2012-01-08 of the NCLT dataset as database and sequence 2012-02-05 as query to compare 4 different setups, including Conv+Conv, SST+Conv, Conv+MST, and SST+MST. Conv+Conv replaces each transformer module with two 1$\times$1 convolution layers without changing the channel numbers. SST+Conv only uses single-scan transformer, while Conv+MST only uses multi-scan transformer. SST+MST represents the holistic SeqOT. As shown in~\tabref{tab:ablation_trans}, the models with transformers consistently outperform the one with only convolution layers. The performance increases significantly if we use MST to fuse spatial and temporal features rather than convolution layers. Besides, The effectiveness of MST is more significant than SST, which means extracting spatial-temporal features with transformer gains more improvement in place recognition compared to only exploiting spatial features.}

% RIE+3TM+GDG and RIE+1TM+GDG outperform the other setups for place recognition. The results show that one transformer block already increases the performance significantly, while more transformer blocks lead to lower efficiency. When using more than 3 transformer blocks, the performance even decreases slightly. More transformer blocks might need more training data and time to obtain good performance. Thus, we only use one transformer block.}

\change{
\begin{table}[t] \color{black}
  \centering
  %\vspace{0.2cm}
  \setlength{\tabcolsep}{3pt}
  \renewcommand\arraystretch{1.3}
  \caption{\change{Ablation study on the transformer modules}}
  \footnotesize{
\begin{tabular}{L{60pt}|C{45pt}|C{35pt}C{35pt}C{36pt}}
%\begin{tabular}{lccc}
\toprule
Network    & Runtime [ms]     & AR@1               & AR@5            & AR@20                 \\ \hline
Conv+Conv    & 67.55         & 0.605               & 0.800            & 0.924                  \\ \hline
SST+Conv      & 88.74              & 0.769             & 0.871             & 0.941                      \\ \hline
Conv+MST   & 77.11         & 0.875         & 0.924        & 0.960         \\ \hline
SST+MST   & 98.32      & \textbf{0.917}  & \textbf{0.947}            & \textbf{0.968}                  
 \\ \bottomrule
\end{tabular}
  }
  \label{tab:ablation_trans}
  \vspace{-0.4cm}
\end{table}
}

%%%%%%%%%%%%%%%%%%%%%%%%%%%%%%%%%%%%%%%%
\subsection{Study on Yaw-Rotation Invariance}
\label{sec:yaw-angle-invariance}

\begin{figure}[t]
  \vspace{0.4cm}
  \centering
  \includegraphics[width=0.79\linewidth]{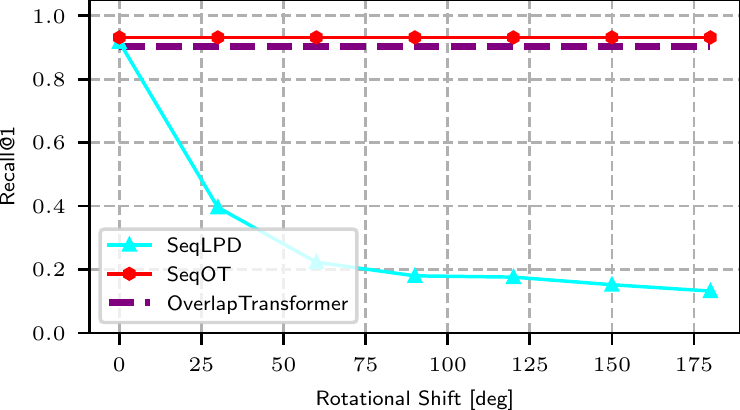}
  \caption{Study on yaw-rotation invariance.}
  \label{fig:yaw_angle_inv_fig}
    \vspace{-0.4cm}
\end{figure}

In this experiment, we validate that our SeqOT is yaw-rotation-invariant against LiDAR scans obtained at the same place but from different viewpoints. We compare our method with SeqLPD conducted on the Haomo dataset. We rotate each query scan of sequence 2-2 along the yaw-axis in steps of 30 degrees and search the places in the same original database. We use Recall@1 as the criterion to evaluate the effect of yaw angle change on our method and SeqLPD. The experimental results are shown in  \figref{fig:yaw_angle_inv_fig}. As can be seen, both OverlapTransformer and SeqOT are not affected by the rotation along the yaw-axis due to our devised yaw-rotation-invariant global descriptors, while the performance of SeqLPD decreases quickly with the yaw angle increasing. This experiment also verifies again that our sequence-enhanced method consistently outperforms the single-scan baseline methods with large viewpoint changes.

%%%%%%%%%%%%%%%%%%%%%%%%%%%%%%%%%%%%%%%%
\subsection{Runtime}
\label{sec:runtime}

\change{We further evaluate the runtime and show our method can achieve online operation. We conduct all experiments on a system with an Intel i7-11700K CPU and an Nvidia RTX 3070 GPU. We run all the models to find top-20 candidates. Here we consider the most extreme case in which all the sub-descriptors are generated from scratch. The holistic SeqOT takes 98.32\,ms for generating the descriptor of each scan and finding the candidates, which is less than 100\,ms thus faster than the sensor frame rate. Besides, the number of parameters of SeqOT is 12.82\,M, which means there is only a little memory consumption.}

\section{Conclusion}
\label{sec:conclusion}

In this paper, we presented a novel end-to-end transformer network for LiDAR-sequence enhanced place recognition. Our approach utilizes Transformer modules at different scales to generate sub-descriptors that fuse the spatial and temporal information provided in sequential LiDAR range images. In the end, a GeM pooling is used to further exploit longer temporal information, fuse the sub-descriptors, and generate a lightweight global descriptor for each sequence. We compared the place recognition performance of our method with the state-of-the-art single-scan and sequence-enhanced methods on four different datasets. The experimental results suggest that our method outperforms the state-of-the-art methods in terms of place recognition performance. The additional zero-shot evaluations on the KITTI and MulRan datasets also show the strong generalization ability of our method. Furthermore, our proposed SeqOT is yaw-rotation-invariant and operates online which can be used for real-world applications. 

%Despite these encouraging results, there are several avenues for future research. First, we want to make our pooling module trainable while using the whole sequence as training input, instead of three consecutive scans separately. We furthermore plan to stack SeqOT and generate more discriminative final-descriptors.

%%%%%%%%%%%%%%%%%%%%%%%%%%%%%%%%%%%%%%%%%%%%%%%%%%%%%%%%%%%%%%%%%%%%%%%%%%%%%%%%
% Only if applicable
%\section*{Acknowledgments}
%We thank XXX for fruitful discussions and for \dots

%% \clearpage
%\bibliographystyle{plain_abbrv}
\bibliographystyle{IEEEtran}

% All new citations should go to new.bib. The file glorified.bib should go
% be the one from the ipb server. After paper or related work has been
% written merge the entries from new.bib to glorified.bib ON THE SERVER,
% replace the glorified.bib in this repository and empty the new.bib
\footnotesize{
\bibliography{ref}}

\begin{IEEEbiography}[{\includegraphics[width=1in,height=1.25in,clip,keepaspectratio]{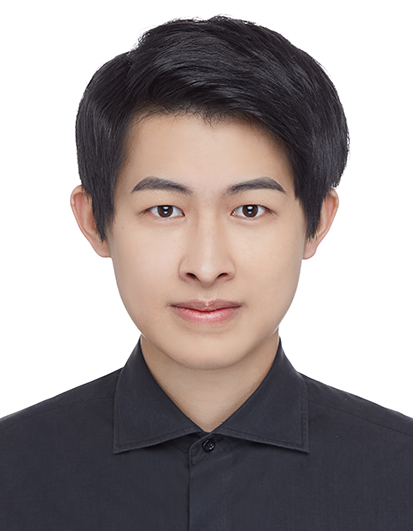}}]{Junyi Ma}
received the B.E degree in vehicle engineering from the Beijing Institute of Technology, Beijing, China, in 2020. He is currently pursuing the Master dgree from the Beijing Institute of Technology. 
His research interests include simultaneous localization and mapping, place recognition, and robotics. He is trying to apply machine learning methods to robotics and use multiple sensor data for enhanced perception capability of robots and intelligent vehicles.
\end{IEEEbiography}
%\vspace{-100 mm} 
\begin{IEEEbiography}[{\includegraphics[width=1in,height=1.25in,clip,keepaspectratio]{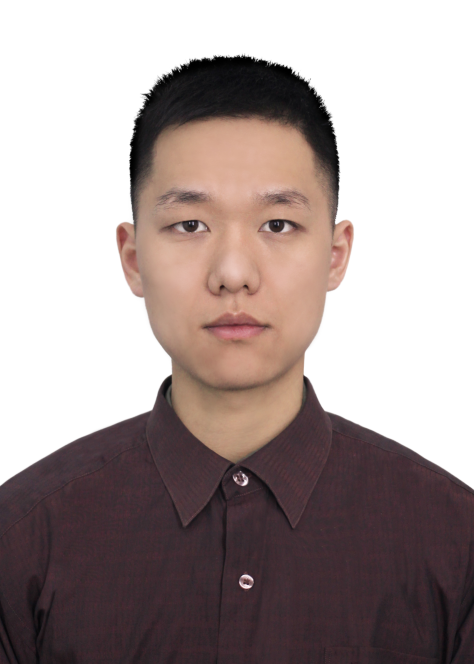}}]{Xieyuanli Chen}
received his Ph.D. degree in Robotics at Photogrammetry and Robotics Laboratory, University of Bonn. He is now also a member of the Technical Committee of RoboCup Rescue Robot League (RRL). He received his Master degree in Robotics in 2017 at the National University of Defense Technology, China. During that time, he was a member of the Organizing Committee of RoboCup Rescue Robot League. He received his Bachelor degree in Electrical Engineering and Automation in 2015 at Hunan University, China. 
\end{IEEEbiography}
\vspace{-150 mm} 
\begin{IEEEbiography}[{\includegraphics[width=1in,height=1.25in,clip,keepaspectratio]{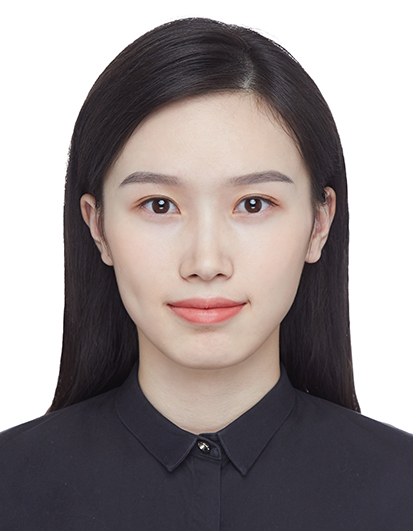}}]{Jingyi Xu}
received the B.E degree in vehicle engineering from the Beijing Institute of Technology, Beijing, China, in 2020. She is currently pursuing the Master dgree from the Beijing Institute of Technology. 
Her research interests include energy management strategies for intelligent vehicles, semantic segmentation, and motion planning of robots.
\end{IEEEbiography}
\vspace{-150 mm} 
\begin{IEEEbiography}[{\includegraphics[width=1in,height=1.25in,clip,keepaspectratio]{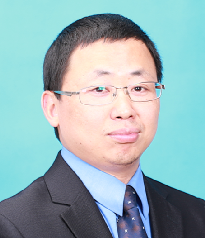}}]{Guangming Xiong}
received the Ph.D. degree in mechanical engineering from the Beijing Institute of Technology, Beijing, China, in 2005. He is currently an Associate Professor with the School of Mechanical Engineering, Beijing Institute of Technology. 
His research interests include intelligent vehicles, mobile robotics, machine vision, and multi-vehicle coordination.
\end{IEEEbiography}

\end{document}